\documentclass{article}

\usepackage{arxiv}

\usepackage[utf8]{inputenc} % allow utf-8 input
\usepackage[T1]{fontenc}    % use 8-bit T1 fonts
\usepackage{hyperref}       % hyperlinks
\usepackage{url}            % simple URL typesetting
\usepackage{booktabs}       % professional-quality tables
\usepackage{amsfonts}       % blackboard math symbols
\usepackage{nicefrac}       % compact symbols for 1/2, etc.
\usepackage{microtype}      % microtypography
\usepackage{lipsum}
\usepackage{graphicx}
\usepackage{cite}
\usepackage{amsmath,amssymb,amsfonts}
\usepackage{algorithmic}
\usepackage{graphicx}
\usepackage{textcomp}
\usepackage{float}
\usepackage{subcaption}
\usepackage{csquotes}
\usepackage{comment}
\graphicspath{ {./images/} }

\title{Uncertainty-aware Remaining Useful Life predictors}

\author{%
  Luca Biggio$^{1,2}$, Alexander Wieland$^1$, Manuel Arias Chao$^{1}$, Iason Kastanis$^2$, Olga Fink$^1$ \\
  $^1$ETH Zürich, Switzerland\\
  $^2$CSEM SA, Switzerland\\
}

\begin{document}
\maketitle
\begin{abstract}
Remaining Useful Life (RUL) estimation is the problem of inferring how long a certain industrial asset can be expected to operate within its defined specifications. Deploying successful RUL prediction methods in real-life applications is a prerequisite for the design of intelligent maintenance strategies with the potential of drastically reducing maintenance costs and machine downtimes. In light of their superior performance in a wide range of engineering fields, Machine Learning (ML) algorithms are natural candidates to tackle the challenges involved in the design of intelligent maintenance systems. In particular, given the potentially catastrophic consequences or substantial costs associated with maintenance decisions that are either too late or too early, it is desirable that ML algorithms provide uncertainty estimates alongside their predictions. However, standard data-driven methods used for uncertainty estimation in RUL problems do not scale well to large datasets or are not sufficiently expressive to model the high-dimensional mapping from raw sensor data to RUL estimates. In this work, we consider Deep Gaussian Processes (DGPs) as possible solutions to the aforementioned limitations. We perform a thorough evaluation and comparison of several variants of DGPs applied to RUL predictions. The performance of the algorithms is evaluated on the N-CMAPSS (New Commercial Modular Aero-Propulsion System Simulation) dataset from NASA for aircraft engines. The results show that the proposed methods are able to provide very accurate RUL predictions along with sensible uncertainty estimates, providing more reliable solutions for (safety-critical) real-life industrial applications.
\end{abstract}

% keywords can be removed

\section{Introduction}\label{sec:introduction}
{Recently,} Predictive Maintenance (PM) methods have been gaining popularity for many different applications. PM aims at predicting the need for maintenance actions based on the information extracted from condition monitoring data describing the health state of the system. Efficient RUL estimation is a key enabler of PM and the application of ML techniques to RUL prediction tasks has been an active research area \cite{l1,l2,l3,l4,l5,l6}.

While the majority of model-based prognostics approaches quantify the associated uncertainty, only few research studies on data-driven RUL prediction algorithms have tackled the challenge of quantifying the level of uncertainty associated with their predictions \cite{fink2020potential, l7}. Uncertainty Quantification (UQ) is crucial in the context of PM because RUL models are used for critical decision-making and, therefore, need to be transparent about the level of uncertainty in their predictions. As a result, the deployment of ML techniques in real-world engineering scenarios cannot prescind from the design of reliable algorithms capable of estimating the level of confidence of their outputs i.e., providing a probability density function over RUL predictions instead of simple point estimates.

%Nonetheless, UQ of data-driven prognostics models has been relatively unexplored so far.\\
While Deep Neural Networks (DNN) have delivered their most prominent achievements in the fields of Computer Vision (CV) and Natural Language Processing (NLP), recent research works have shown their effective use also for prognostics \cite{fink2020potential,b36}. DNNs owe a great part of their success to their large representation power and to their capacity of learning sets of hierarchical features across their multilayer architectures directly from raw data. However, one of the limitations of standard DNN models is that they do not provide an explicit quantification of the uncertainty associated with the predicted RUL. Their effective extension within a Bayesian framework enabling them to perform UQ without sacrificing  their state-of-the-art performances has recently become an active research area in the ML community \cite{wunn,b29,b40,b25}. However, a very limited number of solutions has been proposed for prognostics. Previous works on UQ solutions for purely data-driven prognostics were mainly based on Relevance Vector Machine (RVM) \cite{b37} and Gaussian Process (GP) regression \cite{b38}. GPs, in particular, provide a good adaptability to handle nonlinear, relatively complex regression problems and, compared to standard neural networks, they are based on a well-understood probabilistic formulation. Their flexibility and the availability of open-source software implementations \cite{matlab} have led to a number of interesting applications \cite{b33, b34, b35, b36b,b36c} in prognostics of engineered systems.

Despite their desirable properties in terms of UQ and their elegant theoretical formulation, GPs are affected by two main limitations hindering their application to real-world datasets. First and foremost, they suffer from cubic complexity to the data size.  Specifically, given a dataset of $N$ input-output pairs $(\mathbf{X},\mathbf{y}) = \left\{\mathbf{x}_{i},y_i\right\}_{i=1}^{N}$, the exact calculation of the marginal likelihood involves the computation of the inverse of the $N\times{N}$ kernel matrix $\mathbf{K}_{N N}=k(\mathbf{X}, \mathbf{X})$ which results in a computationally prohibitive $\mathcal{O}\left(N^{3}\right)$ cost. Second, their hypothesis space is completely determined by the choice of the kernel function, which might not be complex enough to model certain types of data. 

% Some light on the problems
However, over the last 20 years, a number of efficient solutions to address the aforementioned limitations and to refine standard GP models have been proposed. In this work, we perform a thorough evaluation of three of such GP extensions: namely, Stochastic Variational Gaussian Processes (SVGPs) \cite{b2,b3}, Deep Gaussian Processes (DGPs) \cite{b4, b5} and Deep Sigma Point Processes (DSPPs) \cite{b6,b7} applied to prognostics. The evaluation is performed on a case study for RUL prediction on aircraft engines using the new Commercial Modular Aero-Propulsion System Simulation (N-CMAPSS) dataset from NASA. The methods used in this work lie in between the GP framework for UQ, well known in prognosis, and the world of DL, trying to combine the strengths of both domains, i.e. the capacity of GPs to yield meaningful uncertainty estimates and the scalability and expressive power of modern DNNs.

%\textbf{Add that these methods are appelaing since they retain the same principles underlying GP but enjoy the advantages of deep nets in terms of representational powers. They therefore represent good comprimeses between a well-known method in prognosis (GP) and the represetational power of deep nets.}

% The contribution
The contribution of this paper is twofold: 1) We demonstrate that DGP models can be successfully employed to predict RUL of complex industrial systems; 2) Our results highlight that while retaining the prediction accuracy, DGPs are able to quantify the associated uncertainty. A comparison of the proposed GP-based models with both a standard and a Bayesian DNN, shows that they result in competitive performance both in terms of accuracy and UQ.

% ... And this is what is coming
The remainder of the paper is organized as follows. Section \ref{sec:related} outlines related work on UQ in data driven prognostics. In Section \ref{sec:method}, the applied methods are described. In Section \ref{sec:case_study}, the case study is introduced and the experiments are explained. In section \ref{sec:results}, the results are presented. Finally, a summary of the work and an outlook are given in Section \ref{sec:conclusion}.

\section{Related work}  \label{sec:related}
%Standard DL techniques applied to RUL estimation range from relatively complex fully-connected networks to CNN or RNN based models. DL techniques have indeed the potential to bypass (or at least to limit) the process of manual feature extraction that is often performed by engineers. This aspect makes them particularly appealing in the context of PM, where the goal is to reduce the impact of human inductive biases and increase the level of automation of maintenance strategies. However, most of the DL approaches proposed in the literature do not take into account the level of uncertainty associated with their predictions, thus posing a strong limit on their deployment. 
%UQ in prognostics

%Uncertainty quantification and management is well developed in the context of model-based prognostics. 
%1) sampling methods and analytical methods for estimating the
%uncertainty in the remaining useful life prediction for prognostics. 
%2) Also Bayesian tracking approaches as Kalman filtering, particle filtering
%\textbf{TODO: Rephrase it and add citations}

%UQ in data-driven prognostics.

\subsection{DL techiques in progostics}
Over the last few years, different types of DNNs have been developed for RUL prediction, ranging from relatively complex fully-connected networks to Convolutional Neural Networks (CNN) \cite{cnn1,cnn2,cnn3} and Recurrent Neural Networks (RNN) \cite{rnn1,rnn2,rnn3}. DL models have shown promising performance in estimating the RUL from sensor data on prognostics benchmark datasets \cite{cmapps,pronostia} using several different network architectures (see \cite{b36} for an extensive review). More sophisticated extensions to the aforementioned standard architectures have also recently been applied to prognosis, including attention mechanisms \cite{b48} and capsule neural networks \cite{ruiz2020novel}. However, a common drawback shared by the majority of the DL models proposed in the literature is that they do not provide uncertainty estimates associated with their predictions, thus, posing a strong limit on their deployment in real-life applications.

\subsection{Bayesian DNNs}
Equipping DL predictions with meaningful uncertainty estimates is a very active research area in the ML community. In the context of Bayesian DL, the central idea is to replace overconfident DL models with Bayesian neural networks, where the weights are treated as random variables. A predictive distribution is then obtained through weight marginalization in such a way that uncertainty in weight space is transferred into probabilistic predictions, instead of mere point estimates. A large part of current research is focused on approximating such predictive distributions, whose exact calculation is typically intractable. Popular methods that follow this approach are, for example, Hamiltonian Monte Carlo \cite{b40}, Laplace approximation \cite{ritter}, expectation-propagation \cite{adams} and Variational Inference \cite{wunn, b29, vi3}. Among these, MC Dropout \cite{b29} has been particularly emerging due to its simple yet effective rationale: by applying the dropout technique at inference time and forward-propagating the input data through the network several times, one can approximate the first two moments of the predictive distribution. More details about MC dropout are reported in an apposite paragraph in Section \ref{sec:method}. An alternative class of methods for UQ is Deep ensembles \cite{b25}, a non-Bayesian technique for estimating uncertainty in DNNs based on training multiple models independently and then aggregating their outputs. These methods provide very competitive results but are significantly computationally expensive. 
\subsection{UQ in data-driven prognostics}
In light of their flexible, probabilistic non-parametric framework, GPs have found several applications in prognosis, e,g, nuclear component degradation \cite{b33}, lithium-ion batteries \cite{b34,b36b,b36c} and bearings \cite{b35}.\\
On the other hand,
despite the increasing efforts in integrating DNNs with effective UQ techniques, very few of the methods mentioned in the previous subsection have been successfully transferred to prognosis tasks. For instance, ensemble approaches were applied for UQ in prognostics in \cite{b28} where, rather than simply training independent models, Bayesian model-averaging was also applied to each model in order to obtain multiple predictions for elements in the ensemble. UQ based on Bayesian neural networks and variational inference were only recently investigated in \cite{b21,b44} with relatively good results in terms of UQ.\\
The goal of this work is to introduce a new class of methods, DGP, that tries to integrate the benefits of DL models into the well-understood Bayesian framework of GP regression. We elaborate more on these approaches in the following sections.

\section{Methods} \label{sec:method}
In this section, we briefly review the basic features of each method employed in our work. A more comprehensive analysis of the GP-based techniques used here can be found, for instance, in \cite{b6}, from which the discussion below is largely inspired. In order to compare our GP-based methods with a strong Bayesian DL baseline, we additionally implement Monte Carlo Dropout (MCD) \cite{b29} and we apply it to the same RUL benchmark dataset. Our choice of MCD is also motivated by the interpretation provided in \cite{b29}, which establishes a connection between MCD and the probabilstic GP introduced in \cite{b4}.
A description of the main principles underlying MCD can be found at the end of this section.

\paragraph{Stochastic Variational Gaussian Processes - SVGPs.}
SVGPs is a popular inducing point method \cite{b8,b9} based on variational inference \cite{b10} that enables the application of the GP framework to big datasets. SVGPs introduce a multivariate Normal variational distribution, $q(\mathbf{u})=\mathcal{N}(\mathbf{m}, \mathbf{S})$, over the inducing variables $\mathbf{u}$. These variables are obtained from a set of inducing points $\mathbf{Z}$ = $\left\{\mathbf{z}_{i}\right\}_{i=1}^{M}$, lying within the same space as $\mathbf{X}$, through the data generating function $f$, i.e. $\mathbf{u} = f(\mathbf{Z})$. The parameters of such a distribution can be estimated through the optimization of the ELBO (evidence lower bound), which can be compactly written as follows:

\begin{equation}\label{ELBO}
\begin{split}
\mathcal{L}_{\mathrm{svgp}}=& \sum_{i=1}^{N}\left\{\log\mathcal{N}\left(y_{i}\mid \mu_{\mathrm{f}}\left(\mathbf{x}_{i}\right), \sigma_{\mathrm{\text {obs}}}^{2}\right)-\frac{\sigma_{\mathrm{f}}\left(\mathbf{x}_{i}\right)^{2}}{2 \sigma_{\text{obs}}^{2}}\right\}-\\&\mathrm{KL}(q(\mathbf{u}) \mid p(\mathbf{u}))
\end{split}
\end{equation}

     where $\sigma^2_{obs}$ is the variance of the Normal likelihood $p(\mathbf{y} \mid f)$, KL denotes the Kullback-Leibler divergence and the two terms $\mu_{\mathrm{f}}$ and $\sigma_{\mathrm{f}}$ indicate the predictive mean and the latent function variance respectively, which have the following form:
    
    \begin{equation}\label{usi}
    \begin{split}
    &    \mu_{\mathrm{f}}\left(\mathbf{x}_{i}\right)=\mathbf{k}_{i}^{T} \mathbf{K}_{M M}^{-1} \boldsymbol{m}\\
    &\sigma_{f}\left(\mathbf{x}_{i}\right)^{2}=\tilde{\mathbf{K}}_{i i}+\mathbf{k}_{i}^{T} \mathbf{K}_{M M}^{-1} \mathbf{S} \mathbf{K}_{M M}^{-1} \mathbf{k}_{i}
    \end{split}    
    \end{equation}
    
    where $\tilde{\mathbf{K}}_{N N}=\mathbf{K}_{N N}-\mathbf{K}_{N M} \mathbf{K}_{M M}^{-1} \mathbf{K}_{M N}$, $\mathbf{k}_{i}=k\left(\mathbf{x}_{i}, \mathbf{Z}\right), \mathbf{K}_{M M}=k(\mathbf{Z}, \mathbf{Z})$ and $\mathrm{K}_{N M}=\mathbf{K}_{M N}^{\mathrm{T}}=k(\mathbf{X}, \mathbf{Z})$.\\
    Given a new test datum $\mathbf{x}^*$, SVGPs yield the following Normal predictive distribution over the corresponding test output $y^*$:
    \begin{equation}\label{predictivesvgp}
    p\left(y_{*} \mid \mathbf{x}_{*}\right)=\mathcal{N}\left(y_{*} \mid \mu_{f}\left(\mathbf{x}_{*}\right), \sigma_{f}\left(\mathbf{x}_{*}\right)^{2}+\sigma_{\mathrm{obs}}^{2}\right)
    \end{equation}
    \\ SVGPs present two main advantages over standard GPs: first, their formulation involves at most the calculation of $\mathbf{K}_{M M}^{-1}$ which results in a significant computational advantage if $M\ll{N}$. Second, the objective in Eq. \ref{ELBO} is written as a sum over single data-points and naturally lends itself to mini-batch training.

\begin{figure}
    \centering
    \includegraphics[scale = 0.5]{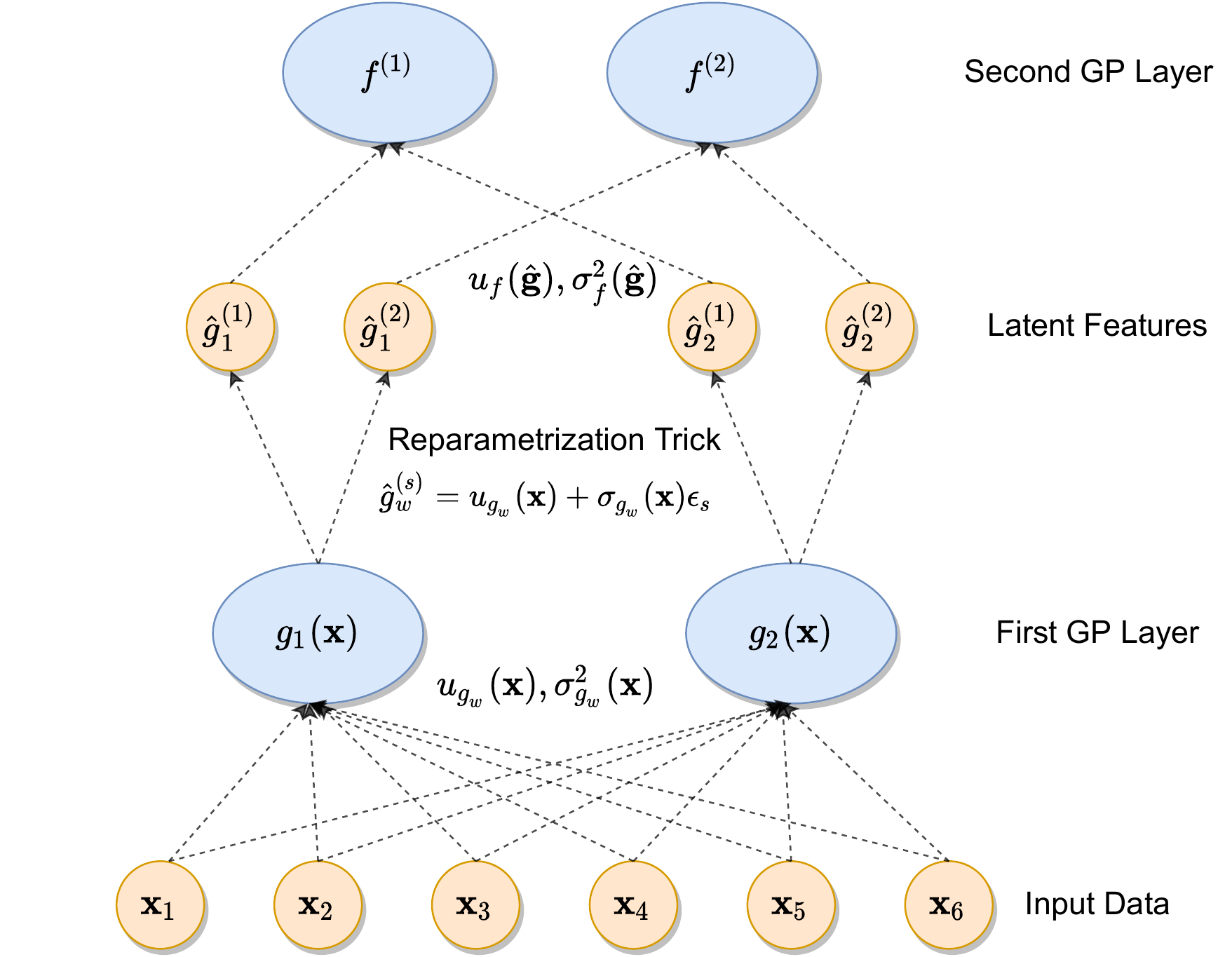}
    \caption{2-layer DGP architecture. The first hidden layer consists of $W=2$ GPs taking as input the data $\mathbf{x}$ to calculate $\mu_{g_{w}}(\mathbf{x})$ and $\sigma_{g_{w}}(\mathbf{x})$ as prescribed by Eq. \ref{usi}. The re-parametrization trick is then used to sample from $\mathcal{N}\left(\mu_{g_{w}}(\mathbf{x}), \sigma_{g_{w}}(\mathbf{x})\right)$ and obtain the features to be fed into the next GP layer.} %The output distribution is described in Eq. \ref{post} and depends on $S$ Gaussians obtained from $S$ sampled feature sets.}
    \label{DeepGP}
\end{figure}

\paragraph{Deep Gaussian Processes - DGPs.}
The classes of functions modelled by standard GP models, including SVGPs, are limited by the expressiveness of the chosen kernel. One possible way to tackle this shortcoming is to use a deep neural network to automatically learn the kernel from data \cite{b11,b12}. However, these approaches often require problem-specific architectures and are prone to overfitting.\\ DGPs consist of hierarchical compositions of GPs and offer a powerful alternative solution to increase the representational power of \enquote{single-layer}-GPs. They retain many of the advantages of shallow GPs and introduce a relatively small number of parameters to optimize compared to standard neural network models. \\In this work, we apply a variant of DGPs recently proposed in \cite{b5} to overcome some of the drawbacks of the original DGP formulation \cite{b4}. This improved model enjoys the same advantages as SVGPs, i.e. it reduces computational complexity by introducing inducing variables for each GP in the hierarchy and supports mini-batch training. More specifically, similarly to SVGPs,  the following ELBO\footnote{Here we consider the case of a 2-layer DGP, for the sake of clarity.} is optimized:

\begin{equation}\label{klkl}
\mathcal{L}_{\mathrm{dsvi}}=\mathbb{E}_{Q}\left[\log p\left(\mathbf{y} \mid \mathbf{f}, \sigma_{\mathrm{obs}}^{2}\right)\right]-\sum\mathrm{KL}    
\end{equation}

where $Q = Q\left(\mathbf{f}, \mathbf{u}_{f}, \ldots, \mathbf{g}_{W}, \mathbf{u}_{g_{W}}\right)$ is a variational distribution depending on each of the GP's latent function values and the corresponding inducing variables. Hidden GP latent functions values are referred to as $\mathbf{g}_{w}$, with $w = 1,..,W$ where $W$ is the number of GPs in the first hidden layer. The KL term in Eq. \ref{klkl} is of the same form as in the SVGP objective and it is summed over all the inducing variables in the DGP architecture, i.e. $\left\{\mathbf{u}_{f}, \ldots, \mathbf{u}_{g_W}\right\}$. The first term in Eq. \ref{klkl} can be written as a sum over data points since sampling from $Q$ can be reduced to sampling from $\left\{q\left(f_{i}\right), \ldots, q\left(g_{i w}\right)\right\}$ where the index $i$ ranges over the number of data points $N$. The resulting method is based on a doubly-stochastic variational inference pipeline since the sampling procedure involves the use of the re-parametrization trick \cite{b13} and the minimization of the factorized objective can be performed with mini-batch training. An illustration of a 2-layer DGP architecture implementing the technique introduced in \cite{b5} is provided in Fig. \ref{DeepGP}.\\
The final predictive distribution can be written as a continuous mixture of Normal distributions:

\begin{equation}\label{second preds}
\mathbb{E}_{\prod_{w=1}^{W} q\left(g_{* w} \mid \mathbf{x}_{*}\right)}\left[\mathcal{N}\left(y_{*} \mid \mu_{f}\left(\mathbf{g}_{*}\right), \sigma_{f}\left(\mathbf{g}_{*}\right)^{2}+\sigma_{\mathrm{obs}}^{2}\right)\right]
\end{equation}
where the expectation is analytically intractable but can be approximated via Monte Carlo samples, resulting in a \textit{finite} mixture of Gaussians.

\paragraph{Deep Sigma Point Processes - DSPPs.}
Despite their many practical successes, variational inference methods often tend to provide overly confident uncertainty estimates \cite{b14, b15}.\\
A recent series of works \cite{b6,b7} aimed to address this limitation by reformulating the variational inference scheme at the basis of SVGPs and DGPs. In particular, the authors note an inconsistency between the ELBO (the objective function to be optimized shown in Eq. \ref{ELBO}) and the predictive distribution to be used at test time (Eq. \ref{predictivesvgp}). More specifically, both quantities are written as functions of two variance terms, one input-dependent, $\sigma_{f}(x)^2$, and one input-independent $\sigma_{obs}^2$. However, these two contributions appear asymmetrically in Eq. \ref{ELBO}. By opportunely modifying the ELBO to fix the aforementioned asymmetry between the objective and the predictive posterior, the authors introduce a new loss function where the two variance terms, $\sigma_{f}(x)^2$ and $\sigma_{obs}^2$, are treated consistently. The new objective is given by:

\begin{equation}
\begin{aligned}
\mathcal{L}_{\mathrm{ppgpr}}=& \sum_{i=1}^{N} \log p\left(y_{i} \mid \mathbf{x}_{i}\right)-\beta_{\mathrm{reg}} \mathrm{KL}(q(\mathbf{u}) \mid p(\mathbf{u})) \\
=& \sum_{i=1}^{N} \log \mathcal{N}\left(y_{i} \mid \mu_{f}\left(\mathbf{x}_{i}\right), \sigma_{f}\left(\mathbf{x}_{i}\right)^{2}+\sigma_{\mathrm{obs}}^{2}\right) \\
&-\beta_{\mathrm{reg}} \operatorname{KL}(q(\mathbf{u}) \mid p(\mathbf{u}))
\end{aligned}
\end{equation}

where $\beta_{\mathrm{reg}}$ acts as a regularization hyperparameter.\\
In \cite{b7}, the authors show that equipping SVGPs with this new objective results in a significant improvement in terms of UQ.\\ In \cite{b6}, the DGP framework proposed in \cite{b5} is combined with the new loss function introduced in \cite{b7}.\\
DSPPs arise from the necessity of overcoming one last theoretical obstacle: the direct application of the objective introduced in \cite{b7} to the DGP predictive distribution in Eq. \ref{second preds} would result in the computation of the logarithm of a continuous mixture of Normal distributions. The approximation of such expectation via Monte Carlo sampling would yield a biased estimator. \\
To cope with this issue, the authors propose to replace the continuous mixture of Gaussians with a parametric (finite) mixture of Gaussians. This procedure is practically implemented by applying an opportune quadrature rule (e.g. the Gauss-Hermite quadrature rule). To better understand this point, let's rewrite Eq. \ref{second preds} as:

\begin{equation}\label{rrr}
\begin{split}
&p\left(y_{i} \mid \mathbf{x}_{i}\right)=\\
&\int d \mathbf{g}_{i} \mathcal{N}\left(y_{i} \mid \mu_{f}\left(\mathbf{g}_{i}\right), \sigma_{f}\left(\mathbf{g}_{i}\right)^{2}+\sigma_{\mathrm{obs}}^{2}\right) \prod_{w=1}^{2} q\left(g_{i w} \mid \mathbf{x}_{i}\right)
\end{split}
\end{equation}

and approximate each distribution inside the product over $w$ as a $S$-component mixture of delta distributions, i.e:
\begin{equation}
\begin{array}{l}
\begin{split}
&\prod_{w=1}^{W} q\left(g_{i w} \mid \mathbf{x}_{i}\right) \rightarrow \\
&\sum_{s=1}^{S} \omega^{(s)} \prod_{w=1}^{W} \delta\left(g_{i w}-\left(\mu_{g_{w}}\left(\mathbf{x}_{i}\right)+\xi_{w}^{(s)} \sigma_{g_{w}}\left(\mathbf{x}_{i}\right)\right)\right)
\end{split}
\end{array}
\end{equation}

where $\left\{\omega^{(s)}\right\}_{s=1}^{S}$ and $\left\{\xi_{w}^{(s)}\right\}_{s=1}^{S}$ are sets of learnable parameters. This transformation allows us to replace
the continuous mixture of Gaussians in Eq. \ref{rrr} with a (parametric) finite mixture by exploiting the properties of the Dirac delta function.

\paragraph{Monte Carlo Dropout - MCD.}
The Dropout technique \cite{b32} is based on randomly dropping units and the corresponding weights from a neural network at training time. As pointed out in the original paper, this procedure results in sampling an exponential number of different \enquote{thinned} networks during training. Predictions are then made by using the entire network including all units and connections. The resulting technique is straightforward to implement and provides an effective strategy to contrast overfitting in DNNs. \\
MCD provides a Bayesian interpretation of the classic Dropout framework and shows that, by enabling dropout at test time, an approximation to a Bayesian Neural Network can be obtained and standard point predictions can be paired to meaningful uncertainty estimates. Furthermore, it can be shown that a standard neural network model with dropout applied before every weight layer represents an approximation to the probabilistic Gaussian Process introduced in \cite{b4}. This observation motivates the analysis of MCD in the context of our work in light of its close relation with GP-based models.

As already mentioned earlier, in Bayesian inference, given a model with parameters $W$ (in the case of MCD, these will be the weights and the biases of the neural network), the final goal is to calculate a predictive posterior distribution $p(\mathbf{y}^*|\mathbf{x}^*, \mathbf{X}, \mathbf{Y})$ for a new data point $(x^*, y^*)$ as 

\begin{equation}
    p(\mathbf{y}^*|\mathbf{x}^*, \mathbf{X}, \mathbf{Y}) = \int p(\mathbf{y}^*|\mathbf{x}^*, W)p(W | \mathbf{X}, \mathbf{Y}) \mathrm{d}W.
    \label{eq:BNN}
\end{equation}

However, the likelihood distribution $p(\mathbf{y}|\mathbf{x}, W)$ is typically a very complicated function of the weights, due to the complex nonlinear mapping implemented by the neural network. This aspect effectively prevents the analytical calculation of the weight posterior and the predictive posterior. \\
The framework of variational inference aims at tackling this problem by introducing an approximation, $q_{\theta}(W)$, of the true posterior, $p(\mathbf{y}^*|\mathbf{x}^*, \mathbf{X}, \mathbf{Y})$,  such that:
 
\begin{equation}
KL\left(q_{\theta}(W) \mid(p(W | \mathbf{X}, \mathbf{Y}))\right)=\int q_{\theta}(W) \log \frac{q_{\theta}(W)}{p(W | \mathbf{X}, \mathbf{Y})} dW.
\end{equation}

is minimized for some optimal variational parameters $\theta^*$. It can be easily shown that this optimization problem is equivalent to the maximization of the so-called evidence lower bound (ELBO), $\mathcal{L}_{VI}(\theta)$:

\begin{equation}
    \mathcal{L}_{VI}(\theta) = \int q_{\theta}(W) \log p(\mathbf{Y}|\mathbf{X}, W)\mathrm{d}W - \mathrm{KL}(q_{\theta}(W)|p(W)).
    \label{VIELBO}
\end{equation}

Now, the main novelty introduced in \cite{b29} is a specific form of the approximate posterior $q_{\theta}(W)$ and a resulting unbiased estimator on Eq. \ref{VIELBO}. More specifically, we consider the following from of $q_{\theta}(W)$:

\begin{align*}
    W_i &= M_i \cdot diag([Z_{i,j}]_{j=1}^{K_i})\\
    Z_{i,j} &\sim \mathrm{Bernoulli}(p_i) \; \forall \;i=1,...,L;\: j=1,...,K_{i-1},
\end{align*}
where $M_i$ and $p$ are the variational parameters, $L$ is the number of layers in the network and $K_i$ is the number of nodes in the $i$-th layer. The parameter $p$ represents the probability of keeping the input and can be interpreted as the opposite of the classical dropout rate. This choice allows us to obtain the following unbiased estimator of the ELBO:

\begin{equation}
\mathcal{L}_{MCD} = \frac{1}{N} \sum_{i=1}^{N} E\left(y_{i}, \hat{y}_{i}\right)+\lambda \sum_{i=1}^{L}\left\|W_{i}\right\|_{2}^{2}
\end{equation}

where $E(y_i, \hat{y}_i)$ refers to arbitrary loss function (e.g. Mean Squared Error (MSE) for regression, Softmax for classification). We can now obtain the mean and the variance of an approximation $q_{\theta}(\mathbf{y}^*|\mathbf{x}^*, \mathbf{X}, \mathbf{Y})$ of the true predicting posterior defined in Eq. \ref{eq:BNN} as follows:

\begin{equation}\label{finalMCD}
\begin{aligned}
&\mathbb{E}_{q_{\theta}(\mathbf{y}^*|\mathbf{x}^*, \mathbf{X}, \mathbf{Y})}\left(\mathbf{y}^{\star}\right)  \approx \frac{1}{T} \sum_{t=1}^{T} f^{W}\left(\mathbf{x}^{\star}\right) \\
&\operatorname{Var}_{q_{\theta}(\mathbf{y}^*|\mathbf{x}^*, \mathbf{X}, \mathbf{Y})}\left(\mathbf{y}^{\star}\right)  \approx \tau^{-1} \mathbf{I}_{D}+\frac{1}{T} \sum_{t=1}^{T}\left(f^{W}\left(\mathbf{x}^{\star}\right)\right)^{T} f^{W}\left(\mathbf{x}^{\star}\right) \\
&-\left(\mathbb{E}_{q_{\theta}(\mathbf{y}^*|\mathbf{x}^*, \mathbf{X}, \mathbf{Y})}\left(\mathbf{y}^{\star}\right)\right)^{T} \mathbb{E}_{q_{\theta}(\mathbf{y}^*|\mathbf{x}^*, \mathbf{X}, \mathbf{Y})}\left(\mathbf{y}^{\star}\right)
\end{aligned}
\end{equation}

where $f^W$ is our neural network model parametrized by its weights $W$, $T$ is the number of samples used for averaging and $\tau$ is the model precision. In practice, the equation above tells us that, in order to retrieve the mean and the variance of the approximate predictive posterior, it is sufficient to forward-propagate the input through the trained network $T$ times. Since dropout is enabled at testing time, each iteration will result in a different network model.\\
Note that the variance calculated as in Eq \ref{finalMCD}, contains two terms: the first term models the intrinsic uncertainty, whereas the second term captures the epistemic uncertainty. Since in Eq. \ref{finalMCD}, the first term does not depend on $\mathbf{x}$, what we are ultimately modelling is homoscedastic noise. In order to make the intrinsic uncertainty term more expressive, we allow $\tau$ to depend on $\mathbf{x}$ and we model it by adding an additional output to the network. This consideration results in the following modified version of the variance expression:

\begin{equation}
\begin{aligned}
&\operatorname{Var}_{q_{\theta}(\mathbf{y}^*|\mathbf{x}^*, \mathbf{X}, \mathbf{Y})}\left(\mathbf{y}^{\star}\right)  \approx \frac{1}{T} \sum_{t=1}^{T}\tau^{-1}(\mathbf{x}^{\star})+\left(f^{W}\left(\mathbf{x}^{\star}\right)\right)^{T} f^{W}\left(\mathbf{x}^{\star}\right) \\
&-\left(\mathbb{E}_{q_{\theta}(\mathbf{y}^*|\mathbf{x}^*, \mathbf{X}, \mathbf{Y})}\left(\mathbf{y}^{\star}\right)\right)^{T} \mathbb{E}_{q_{\theta}(\mathbf{y}^*|\mathbf{x}^*, \mathbf{X}, \mathbf{Y})}\left(\mathbf{y}^{\star}\right)
\end{aligned}
\end{equation}

The expression of the variance reported above accounts for the eventual heteroscedasticity of the data noise.

\section{Case Study} \label{sec:case_study}
\subsection{Dataset}
\begin{figure}[ht]
    \centering
    \includegraphics[scale = 0.4]{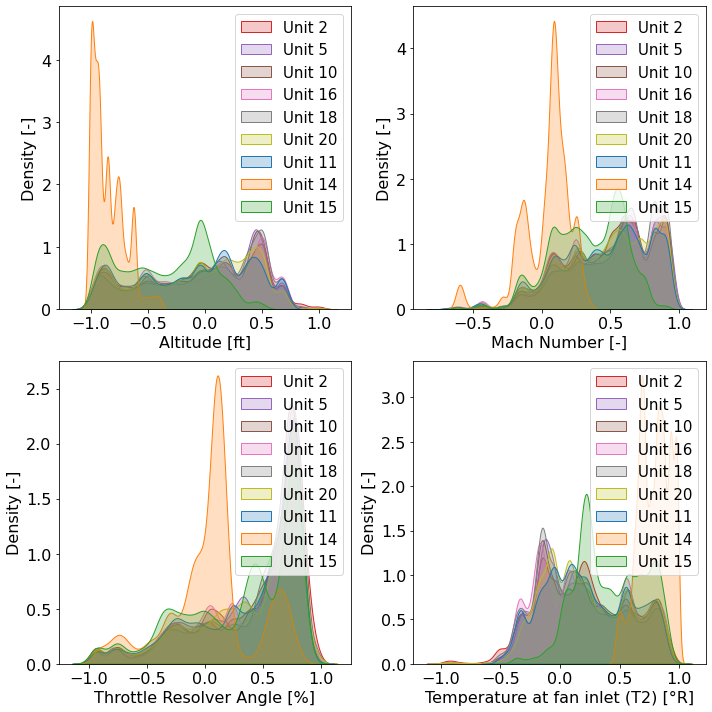}
    \caption{Kernel density estimations of the flight envelopes given by recordings of altitude, flight Mach number, throttle-resolver angle, and total temperature at the fan inlet for complete run-to-failure trajectories of six training units (black) and three test units i.e., Unit 11 (blue), Unit 14 (orange) and Unit 15 (green).}
    \label{kde}
\end{figure}

We evaluate and compare the UQ capabilities of the selected ML techniques on the new C-MAPSS dataset that provides full degradation trajectories of a fleet of nine large turbofan engines under real flight conditions \cite{b16}. Concretely, the flight data cover climb, cruise and descend flight conditions corresponding to different commercial flight routes. The degradation trajectories are given in the form of multivariate time-series of sensor readings. Figure \ref{kde} shows the distribution of the flight envelopes for six training units and three test units. It is worth noting that test unit 14 has an operation distribution that is significantly different from the training units. Concretely, test unit 14 operates shorter and lower altitude flights compared to other units. The training dataset contains, thereby, flight profiles that are not fully representative of the test conditions of this unit. Two distinctive failure modes are present in the development dataset: a high pressure turbine (HPT) efficiency degradation and a more complex failure mode that affects the low pressure turbine (LPT) efficiency and flow in combination with the high pressure turbine (HPT) efficiency degradation. Test units (i.e., Units 11, 14, \& 15) are subjected to the latter complex failure mode. The sampling rate of the data is 0.1 Hz resulting in a total size of the dataset of 0.53M samples for model development and 0.12M samples for testing. More details about the generation process can be found in \cite{b16}.

\subsection{Problem Formulation} \label{sec:Formulation}
Given are multivariate time-series of condition monitoring sensor readings and physics-inferred process features $X_{i} = [x_{i}^{(1)}, \dots, x_{i}^{(n_i)}]^T \in R^{m}$ and their corresponding RUL i.e., $Y_i=[y_i^{(1)},\dots, y_i^{(n_i})]^T$ from a fleet of six units (i.e., $N_{train}$= 6). The length of the input feature vector for the \textit{i-th} unit is given by $n_{i}$; which differs from unit to unit. The total combined length of the available data 
set is $N={\sum_{i=1}^{N_{train}}n_i}$ and the dimension of the input features is 41 (i.e., $m=41$). More compactly, we denote the available dataset as $\mathcal{D} =\{X_{i}, Y_i\}_{i=1}^{N_{train}}$. Given this set-up, the task is to obtain a predictive model that provides a reliable RUL estimate ($\mathbf{\hat{Y}}$) with UQ on a test dataset of $M=3$ units $\mathcal{D}_{T*}=\{X_{s_j*}\}_{j=1}^{M}$.

\subsection{Evaluation Metrics} \label{sec:Metrics}
We evaluate the prediction accuracy and uncertainties obtained by the proposed techniques in terms of the standard RMSE and negative log-likelihood (NLL). In addition, we also incorporate the  $\alpha-\lambda$ metric; which is commonly used in prognostics analysis \cite{b31} and is defined as: 
\begin{equation}
    \alpha - \lambda = 
    \begin{cases}
      1, & \text{if}\ (1-\alpha)\lambda^* \leq \lambda_p \leq (1+\alpha)\lambda^*  \\
      0, & \text{otherwise}
    \end{cases}
\end{equation}
where $\lambda^*$ is the ground truth, $\lambda_p$ the prediction. Therefore, the $\alpha-\lambda$ metric measures if the prediction accuracy of a RUL model is within a $\alpha \; \%$ error at specific time instances during the relative lifetime $\lambda$ of the system. \\
For an arbitrary chosen accuracy $\alpha$, the metric can be evaluated and averaged over the whole trajectory with $N$ samples:
\begin{equation}
    \overline{\alpha - \lambda} = \frac{1}{N}\sum_{n=0}^N (\alpha - \lambda)_n.
\end{equation}

However, this evaluation metric takes only single-value predictions into account, neglecting the uncertainty associated with them. In order to account for predictive uncertainty, a probabilistic version of the standard $\alpha-\lambda$ is used \cite{b49}. Given the variance obtained from the model, we can fit a Gaussian distribution to each output and calculate the probability for a given prediction of being inside the boundary $\alpha$. 
For a generic Gaussian distribution $\mathcal{N}(\mu,\sigma)$ we define the cumulative distribution function as
\begin{equation}
    \textbf{F}(x,\mu,\sigma)  = \Phi \left( \frac{x-\mu}{\sigma}\right) = \frac{1}{2} \left[ 1+ \mathrm{erf} \left(  \frac{x-\mu}{\sigma\sqrt{2}}\right)\right].
\end{equation} 
This allows us to define the probabilistic $\alpha-\lambda$ for a single prediction $\mathcal{N}(\mu,\sigma)$ as
\begin{equation}
    \mathbf{P}_{\alpha-\lambda} = \textbf{F}((1+\alpha)\lambda^*,\mu,\sigma) - \textbf{F}((1-\alpha)\lambda^*,\mu,\sigma).
\end{equation}

Again, for an arbitrary chosen accuracy $\alpha$, the metric can be evaluated and averaged over the whole trajectory with $N$ samples:
\begin{equation}
    \overline{ \mathbf{P}_{\alpha-\lambda}} = \frac{1}{N} \sum_{n=0}^N P_{(\alpha-\lambda)_n}.
\end{equation}
For both cases we set the $\alpha$ value equal to $20\%$ as commonly done in the literature.
% RMSE
% NLL
% and $\alpha-\lambda$ 

%where $m_*$ denotes the total number of test data samples, $\Delta^{(j)}$ is the difference between the estimated and the real RUL of the $j$ sample (i.e., $y^{(j)}-\hat{y}^{(j)}$) and $\alpha$ ... 
% 
\subsection{Model Architectures} \label{sec:Metrics}
\textbf{GP models}. For the SVGP model, we performed a hyperparameter grid-search over the number of inducing points $I \in \{200,400,800\}$. We consider the NLL on the validation set as our model-selection metric. The lowest validation NLL is reached with $I = 800$.\\
The DGP model uses a single hidden layer with a skip-connection enhancing the input, and $W = 4$ hidden GPs. We perform a hyperparameter grid-search over the number of inducing points $I\in\{50,100,200\}$. The lowest NLL on the validation set is achieved for $I = 100$.\\
For the DSPP, we perform a hyperparameter grid-search over the number of inducing points $I\in\{50,100,200\}$, the width of the hidden layer $W\in\{2,3\}$ and the number of quadrature sites $Q\in\{5,8,10,15,20\}$. The lowest NLL on the validation set is reached with $I = 100$, $W=2$ and $Q = 15$.

\textbf{MCD model}. For the MCD approach, we perform a grid-search over the hyperparameter space characterized by $L \in \{2,3,4,5\}$ hidden linear layers with $H_f \in \{50, 65, 80, 100, 150, 200\}$ hidden units each. The dropout rate is searched over a log-range of 12 possible values within the interval $[0.01,1]$. A ReLU function is used after each hidden layer. The lowest NLL on the validation set is achieved for $L=5$, $H_f=200$ and $p=0.4642$.

\textbf{FFNN model}. For the FNN model, we perform a grid-search over the hyperparameter space characterized by $L \in \{2,3,4,5\}$ hidden layers with $H_f \in \{50, 65,80, 100,150, 200\}$ hidden units each. A ReLU function is used after each hidden layer and a constant dropout rate of $p=0.15$. In this case, we consider the RMSE on the validation set as our evaluation metric. The lowest RMSE value is reached with $L=5$ and $H_f=65$.

The batch size is set to 2000 samples using the Adam optimizer with a learning rate of $10^{-3}$ for all the above models.

%In order to have a clear baseline, we implement two deep learning models: a standard deep fully connected feed-forward neural network (FFNN) and a one-dimensional convolutional neural network (1-d CNN). The FFNN model is characterized by  $L \in \{2,3,4,5\}$ hidden linear layers, with $H_f \in \{50, 100, 200\}$ hidden units each. The search space of the 1-d CNN model consists of $C\in\{2,3,4\}$ convolutional modules, each with $F\in\{10, 20,30\}$ filters of size $K\in\{10,20\}$. A fully connected linear layer with $H_c\in \{50, 100\}$ units is used after the convolutional blocks. ReLU activation functions are used in both models.

% All our GP models are implemented using the open-source library GpyTorch \citep{gpytorch}.

\section{Results} \label{sec:results}
In this section, we apply the methods described above to the N-CMAPPS dataset in order to predict the RUL of the three test units (i.e., units 11, 14 and 15) and quantify the uncertainty of the predictions. We report the results provided by all the methods listed above. All the GP-based methods are equipped with the new objective introduced in \cite{b7}\footnote{In particular, in the case of DGPs, we compute a biased estimator of the continuous mixture of Gaussians obtained by applying Monte Carlo sampling.} since we found empirically that, in agreement with the results of \cite{b7}, using the ELBO has a negative impact on the UQ quality for both SVGP and DGPs. All the considered algorithms are implemented using Pytorch. For the GP-based models, we used the open-source library GpyTorch \cite{b19}.

\subsection{Performance analysis.} 
In this section, we compare the prediction accuracy of the considered models in terms of the probabilistic negative log-likelihood (NLL), RMSE, ${\alpha-\lambda}$ and $\mathbf{P}_{\alpha-\lambda}$. The results are shown in Tab. \ref{rmse}. The table shows that  DSPPs provide the best NLL results, whereas MCD only slightly outperforms DGPs and DSSPs in terms of RMSE. While the performances of DSPPs and MCD agree in terms of ${\alpha-\lambda}$ , DSSPs outperforms all the other methods on the $\mathbf{P}_{\alpha-\lambda}$ metric. All the results reported in \ref{rmse} are obtained on the held-out test dataset.

%\textbf{1-d CNN network architecture}. The architecture of the CNN neural network used in this research comprises five convolutional layers with filters of size $10 \times n$, where $n$ denotes the size of the input space $X$ and can vary depending on the considered solution strategy (see Table \ref{tb:input}). The first four convolutions have ten channels and the last convolution has only one filter. Zero padding is used to keep the feature map through the network. The resulting 2D feature map is flattened and the network ends with a 50-way fully connected layer followed by a linear output neuron. The network incorporates dropout as regularization and use \textit{ReLu} as the activation function. The network has 24k trainable parameters ($\mathcal{H}$). The final architecture is the result of a conductig a grid-search over these hyperparameters: number of hidden layers stages [1-4], stages [1-4], channels size [10, 20, 30], filter size [10, 20], activation function type [\textit{tanh}, \textit{relu}], number of neurons at the last hidden layer [50, 100] and window size of the siding window [20, 50, 200].

\begin{table}[H]
\begin{center}
\begin{tabular}{|l|c|c|c|c|} \hline
\multicolumn{5}{|c|}{\textbf{Gaussian Processes}}                     \\ \hline \hline
\textbf{Models}     & \textbf{NLL}  &\textbf{RMSE}   & $\alpha -\lambda$ & $\mathbf{P}_{\alpha-\lambda}$\\ \hline
SVGP                & 3.50          & 8.70           & 0.43              & 0.36                          \\ \hline
DGP                 & 3.24          &7.37   & 0.49              & 0.46                          \\ \hline
DSPP   & \textbf{3.10} & 7.38           & \textbf{0.56}     & \textbf{0.53}                 \\ \hline \hline
\multicolumn{5}{|c|}{\textbf{Deep Neural Networks}}                    \\ \hline \hline
\textbf{Models}     & \textbf{NLL}  &\textbf{RMSE}   & $\alpha -\lambda$ & $\mathbf{P}_{\alpha-\lambda}$ \\ \hline
MCD                 &    4.26       & \textbf{7.31}           & \textbf{0.56}     & 0.48                          \\ \hline
FFNN                & -             & 7.71           & 0.55              & -                             \\ \hline
\end{tabular}
\end{center}
\caption{Comparison of SVGP, DGP, DSPP, MCD, and FNN in terms of negative log-likelihood (NLL), RMSE, $\alpha -\lambda$ and $\mathbf{P}_{\alpha-\lambda}$  on the test data.}\label{rmse}
\end{table}

%The above results clearly show that the RMSE values provided by our three GP-models are compatible with the result given by the neural network models. In terms of NLL, the proposed GP-based models significantly outperform MCD. %In the case of DSPP, the obtained RMSE is slightly below the neural network one. 

\subsection{UQ Analysis} 
\begin{figure}[H]
    \centering
    \includegraphics[scale = 0.4]{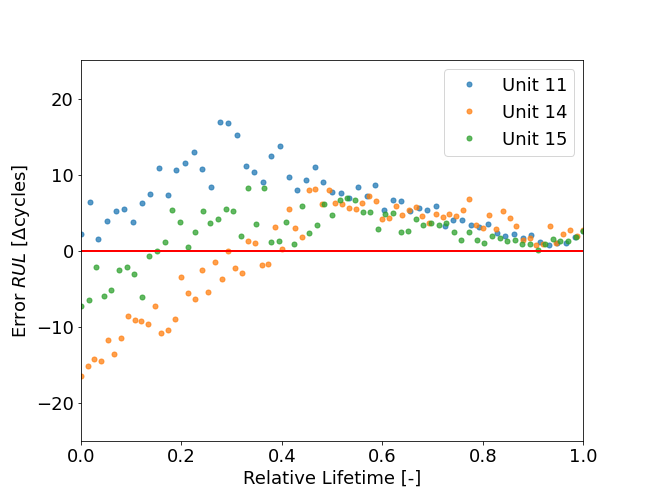}
    \caption{Prediction error of the FFNN models as a function of the relative life time.}
    \label{nonprob}
\end{figure}

\paragraph{Visualizations}
In this paragraph, we provide some visualizations to demonstrate the UQ performance of the proposed methods. Since all our GP-based models provide very similar confidence bounds, we report only the results obtained by the DSPP model.\\
We start our analysis by showing the test prediction error of the considered FFNN model. The results are shown in Fig. \ref{nonprob}.

\begin{figure*} % for sub figures over two columns in 
\centering
\subfloat{\includegraphics[scale=0.35]{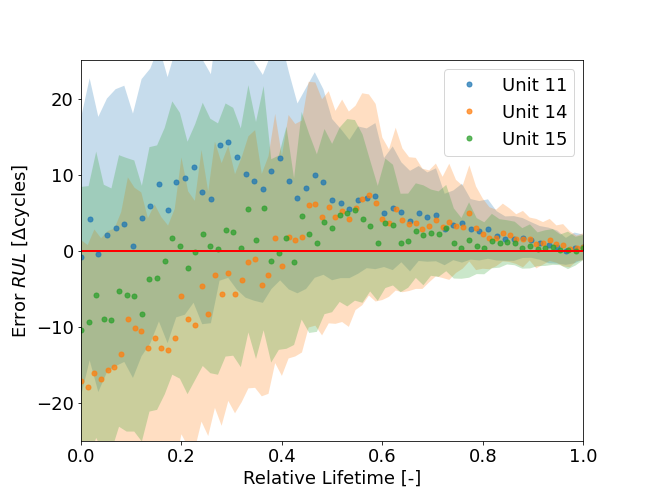}}
\hfill
\subfloat{\includegraphics[scale=0.35]{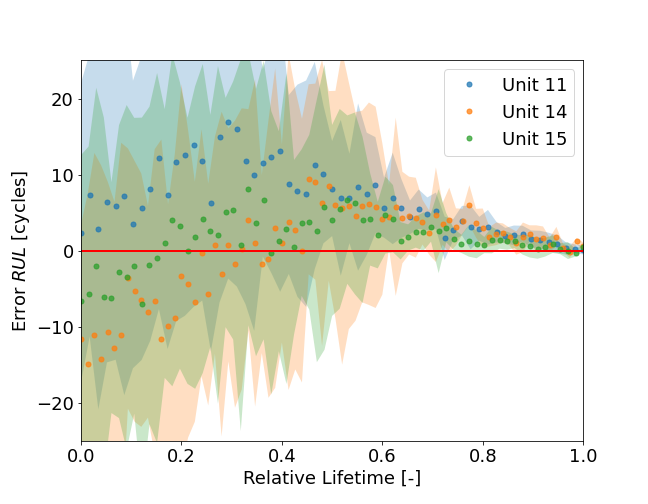}}
\caption{Predicted RUL estimation (dot) including uncertainty (shaded area $\pm 2\sigma$) over the relative time for each test units i.e., Unit 11 (blue), Unit 14 (orange) and Unit 15 (green).}
\label{Fig4}
\end{figure*}

As expected, the predictions tend to align to the ground truth towards the end of the units' lifetime. However, the network is overly confident in its RUL estimates even in the case they significantly diverge from the ground truth (first predictions are far away from the ground truth). \\
On the contrary, as shown in Fig. \ref{Fig4}, the predictions provided by MCD (left) and DSPP (right) are supported by meaningful uncertainty estimates. In both cases, the confidence bounds show an important desirable characteristic for RUL models. The values of the predictive variance decrease over time. This is physically meaningful since predictions  are much more uncertain when the system is far away from its end of life. As a result, the confidence bounds associated with early operating times are significantly larger that those corresponding to the machine's end of life. Such a property has very important practical implications since it enables to design risk-aware maintenance strategies. 
%vspace{-5mm}

%\begin{figure*} % for sub figures over two columns in 
%\centering
%\subfloat{\includegraphics[width=\columnwidth]{best_model.png}}
%\hfill
%\subfloat{\includegraphics[width=\columnwidth]{DSPP_best.png}}
%\caption{Predicted vs. ground-truth RUL Error values for units 11,14,15 for our MCD (left) and a our DSPP model (right). $\pm 2\sigma$ confidence intervals are represented.}
%\label{Fig4}
%\end{figure*}

%\begin{figure}[H]
%    \centering
%    \includegraphics[width=0.45\linewidth]{Figures/unc2 (2).png}
%    \caption{Evolution of the standard deviation over time for test units 11,14,15.}
%    \label{Fig5}
%\end{figure}

%    \textcolor{green}{Here I would show some plots of the results of UQ provided by one or all our models. These can be both at the time-step or at the cycle level. Additionally, one could also show a plot where the variance descreases over time, as expected from physical considerations. This would be in contrast with NNs or CNN deterministic models where outputs are only unreliable point estimates.}

\begin{figure}[ht]
    \centering
    \includegraphics[scale=0.4]{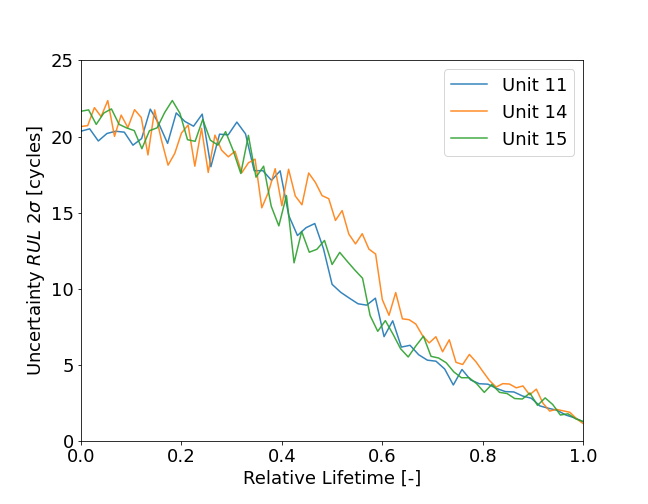} 
    \caption{Evolution of predictive uncertainty provided by DSPP over the relative time for each test units i.e., Unit 11 (blue), Unit 14 (orange) and Unit 15 (green).}
    \label{fig:UQ_per_unit}
\end{figure}

\paragraph{Robustness to a shift between training and testing distribution}
As discussed in Section \ref{sec:case_study}, test unit 14 operates at shorter and lower altitude flights compared to the training units. This aspect results in a distributional shift between training and testing distributions, thus, challenging the generalization capabilities of the proposed method.
In this section, we evaluate whether the obtained UQ models are robust under the aforementioned distributional shift. In particular, we are interested in assessing whether the uncertainty estimations exhibit higher uncertainty on inputs that are far away from the training data distribution.\\
Figure \ref{fig:UQ_per_unit} shows the evolution of the predictive uncertainty (i.e.,  $\pm 2\sigma$) provided by the DSSP model over time for each unit. While at the very first cycles, the level of uncertainty is very high for all of the units (due to the inherent indeterminacy of estimating the RUL when the machines operate under nominal conditions), the RUL predictions for Unit 14 show a larger uncertainty compared to the test units 11 and 15 at later cycles closer to the end of life, when the signs of a fault are increasingly more apparent. Therefore, the confidence bounds of the proposed methods show another important desirable characteristic for RUL models: the values of the predictive variance exhibit a higher uncertainty on inputs that are far away from training data.

\section{Conclusion}\label{sec:conclusion}
In this work, we analyzed a number of methods capable of modelling the uncertainty associated with their predictions. In particular, we have focused on the problem of RUL estimation, i.e. predicting the remaining useful lifetime of an industrial asset of interest. In light of the safety critical nature of this task, UQ is critical in order to allow the deployment of reliable and transparent machine learning algorithms in such real-life industrial applications. The considered methods combine the strengths of neural networks and GPs: they combine the expressive power and scalability of neural networks with the probabilistic nature of GPs. \\
Overall, our results demonstrate that the best performing models are DSPP and MCD: DSSP achieves the highest NLL score while the MCD achieves the best RMSE score, while both of them outperform all the other techniques in terms of ${\alpha-\lambda}$ and $P_{\alpha-\lambda}$. Furthermore, our visualizations show that the confidence bounds provided by the considered models are meaningful: uncertainty decreases as the system approaches the end of life and it is higher for units whose operating conditions differ significantly from those of the training units. These aspects are in strong contrast to the behaviour of a standard deep neural network model which does not take uncertainty into account in its predictions and solely returns point estimates. Last but not least, contrary to the standard GP models, all the proposed methods are characterized by favourable scaling properties and can be applied to large training datasets. \\
In the future, we will focus on two main aspects. First, we would like to extent the proposed methods so that they can better capture the temporal correlations present in time-series data. We expect such a modifications to have a positive impact on the final performance. However, although this is relatively straightforward for MCD (it amounts on replacing the current fully-connected architecture with a one-dimensional convolutional neural network), it is not trivial how to implement it for the GP-based models. Second, we would like to investigate more recent Bayesian DNNs and compare them to the methods proposed in this work. We leave the aforementioned open research directions to future work.

%In this work, we have shown that modern GP models can be successfully applied to the domain of PM of industrial assets. 
%In this work, we provide the first evidence that extensions of GP models, in particular deep GP, can be successfully applied to predict the remaining useful lifetime of industrial assets. The extensions of GP models have helped to make GPs more scalable and expressive.\\
%Our experiments show that the application of such techniques to the new C-MAPSS dataset results in predictive performances close to or superior than those obtained by two strong DL baselines. Nevertheless, the proposed GP-models are able to provide physically meaningful uncertainty estimates alongside their RUL estimates. \\As shown in our visualizations, this aspect is in stark contrast to the behaviour of a standard deep neural network model which does not take uncertainty into account in its predictions.\\ In light of the results obtained, we believe that the models analyzed in this work could be successfully employed in several real-world scenarios, especially in those cases where overly confident predictions might result in catastrophic consequences. 


\begin{thebibliography}{00}

\bibitem{b1} Rasmussen, Carl Edward. "Gaussian Processes in Machine Learning." Advanced Lectures on Machine Learning: 63.

\bibitem{b2}Hensman, James, Alexander G. Matthews, and Zoubin Ghahramani. "Scalable variational Gaussian process classification." Proceedings of Machine Learning Research 38 (2015): 351-360.

\bibitem{b3}Hensman, James, Nicolo Fusi, and Neil D. Lawrence. "Gaussian Processes for Big Data." Uncertainty in Artificial Intelligence. 2013.

\bibitem{b4} Damianou, Andreas, and Neil Lawrence. "Deep gaussian processes." Artificial Intelligence and Statistics. 2013.

\bibitem{b5} Salimbeni, Hugh, and Marc Deisenroth. "Doubly stochastic variational inference for deep Gaussian processes." Advances in Neural Information Processing Systems. 2017.

\bibitem{b6} Jankowiak, Martin, Geoff Pleiss, and Jacob R. Gardner. "Deep Sigma Point Processes." arXiv preprint arXiv:2002.09112 (2020).

\bibitem{b7} Jankowiak, Martin, Geoff Pleiss, and Jacob R. Gardner. "Sparse gaussian process regression beyond variational inference." arXiv preprint arXiv:1910.07123 (2019).

\bibitem{b8} Snelson, Edward, and Zoubin Ghahramani. "Sparse Gaussian processes using pseudo-inputs." Advances in neural information processing systems. 2006.

\bibitem{b9} Titsias, Michalis. "Variational learning of inducing variables in sparse Gaussian processes." Artificial Intelligence and Statistics. 2009.

\bibitem{b10} Blei, David M., Alp Kucukelbir, and Jon D. McAuliffe. "Variational inference: A review for statisticians." Journal of the American statistical Association 112.518 (2017): 859-877.

\bibitem{b11} Wilson, Andrew Gordon, et al. "Deep kernel learning." Artificial intelligence and statistics. 2016.

\bibitem{b12}Wilson, Andrew G., et al. "Stochastic variational deep kernel learning." Advances in Neural Information Processing Systems. 2016.

\bibitem{b13} Kingma, Diederik P., and Max Welling. "Auto-Encoding Variational Bayes." stat 1050 (2014): 1.

\bibitem{b14} Turner, Richard E., Pietro Berkes, and Maneesh Sahani. "Two problems with variational expectation maximisation for time-series models." Inference and Estimation in Probabilistic Time-Series Models.

\bibitem{b15} Bauer, Matthias, Mark van der Wilk, and Carl Edward Rasmussen. "Understanding probabilistic sparse Gaussian process approximations." Advances in neural information processing systems. 2016.

\bibitem{b16} {Arias Chao}, Manuel and Kulkarni, Chetan and Goebel, Kai and Fink, Olga,  "Aircraft Engine Run-to-Failure Dataset under Real Flight Conditions for Prognostics and Diagnostics", \text{Data}, Vol. 6 n. 1, p.5 \textit{2021}.

\bibitem{b17} Arias Chao, Manuel, Kulkarni, Chetan and Goebel, Kai and Fink, Olga. "Fusing physics-based and deep learning models for prognostics." arXiv preprint arXiv:2003.00732 (2020).

\bibitem{b18} Liu, Yuan, et al. "User's guide for the commercial modular aero-propulsion system simulation (c-mapss): Version 2." (2012).

\bibitem{b19} Gardner, Jacob, et al. "Gpytorch: Blackbox matrix-matrix gaussian process inference with gpu acceleration." Advances in Neural Information Processing Systems. 2018.

\bibitem{b20} Sankararaman, Shankar, Goebel, Kai, "Why is the Remaining Useful Life Prediction Uncertain?" PHM 2013 - Proceedings of the Annual Conference of the Prognostics and Health Management Society. 2013.

\bibitem{b21} Peng, Weiwen, et al. "Bayesian Deep-Learning-Based Health Prognostics Toward Prognostics Uncertainty" IEEE Transactions on Industrial Electronics, Vol. 67, No. 3, 2020.

\bibitem{b22} Daigle, Matthew, et al. "A Comparison of Filter-based Approaches for Model-based Prognostics" IEEE AC, 2012

\bibitem{cnn1}Li, Xiang, Qian Ding, and Jian-Qiao Sun. "Remaining useful life estimation in prognostics using deep convolution neural networks." Reliability Engineering and System Safety 172 (2018): 1-11.

\bibitem{cnn2}Wen, Long, Yan Dong, and Liang Gao. "A new ensemble residual convolutional neural network for remaining useful life estimation." Math. Biosci. Eng 16.2 (2019): 862-880.

\bibitem{cnn3}Ren, Lei, et al. "Prediction of bearing remaining useful life with deep convolution neural network." IEEE Access 6 (2018): 13041-13049.

\bibitem{rnn1}Chen, Jinglong, et al. "Gated recurrent unit based recurrent neural network for remaining useful life prediction of nonlinear deterioration process." Reliability Engineering and System Safety 185 (2019): 372-382.

\bibitem{rnn2}Wu, Yuting, et al. "Remaining useful life estimation of engineered systems using vanilla LSTM neural networks." Neurocomputing 275 (2018): 167-179.

\bibitem{rnn3}Wu, Jun \& Hua, Kui \& Yiwei, Cheng \& Zhu, Haiping \& Shao, Xinyu \& Wang, Yuan. (2019). Data-driven remaining useful life prediction via multiple sensor signals and deep long short-term memory neural network. ISA Transactions. 97. 10.1016/j.isatra.2019.07.004. 

\bibitem{cmapps}Saxena, Abhinav \& Goebel, Kai. (2008). C-MAPSS data set. NASA Ames Prognostics Data Repository. 

\bibitem{pronostia}Nectoux, Patrick \& Gouriveau, Rafael \& Medjaher, Kamal \& Ramasso, Emmanuel \& Chebel-Morello, Brigitte \& Zerhouni, Noureddine \& Varnier, Christophe. (2012). PRONOSTIA: An experimental platform for bearings accelerated degradation tests. Conference on Prognostics and Health Management.. 1-8. 

\bibitem{b23} Ellefsen, André Listou, et al. "A Comprehensive Survey of Prognostics and Health Management Based on Deep Learning for Autonomous Ships"  IEEE Transactions on Reliability, Vol. 68, No. 2, 2019.

\bibitem{wunn} Blundell, C., Cornebise, J., Kavukcuoglu, K., $\&$ Wierstra, D. (2015, June). Weight uncertainty in neural network. In International Conference on Machine Learning (pp. 1613-1622). PMLR.


\bibitem{b24} Deutsch, Jason, He, David. "Using Deep Learning-Based Approach to Predict Remaining Useful Life of Rotating Components" IEEE Transactions on Systems, Man. and Cybernetics: Systems. Vol. 48, No. 1, 2018. 

\bibitem{b25} Lakshminarayanan, Balaji, Pritzel, Alexander, Blundell, Charles, " Simple and Scalable Predictive Uncertainty Estimation using Deep Ensembles" Advances in Neural  Information Processing Systems, 2017.

\bibitem{b26} He, Bobby, Lakshminarayanan, Balaji, Teh, Yee Whye, "Bayesian Deep Ensembles via the Neural Tangent Kernel" 

\bibitem{b27} Liao, Y., Zhang, L., Liu, C., "Uncertainty prediction of remaining useful life using long short-term memory network based on bootstrap method". \textit{2018 IEEE International Conference on Prognostics and Health Management (ICPHM)} p. 1-8. \textit{2018}

\bibitem{b28} Deng, Yingjun, et al. "Controlling the accuracy and uncertainty trade-off in RUL prediction with a
surrogate Wiener propagation model" Reliability Engineering and System Safety 196. 2020.

\bibitem{b29} Gal, Yarin, Ghahramani, Zoubin, "Dropout as a Bayesian Approximation: Representing Model Uncertainty in Deep Learning" Procedings oif the 33rd ICML, Vol. 48, 2016. 

\bibitem{b30} Kendall, Alex, Gal, Yarin, "What Uncertainties Do We Need in Bayesian Deep Learning for Computer Vision?" 31st Conference on NIPS, 2017.¨

\bibitem{b31} Saxena et al. "Metrics for Evaluating Performance of Prognostic Techniques"

\bibitem{b32} Nitish Srivastava and Geoffrey Hinton and Alex Krizhevsky and Ilya Sutskever and Ruslan Salakhutdinov, "Dropout: A Simple Way to Prevent Neural Networks from Overfitting", \textit{Journal of Machine Learning Research}, Vol. 14, No. 56, p. 1929-1958, \textbf{2014}

\bibitem{b33} Piero Baraldi, Francesca Mangili, Enrico Zio, "A prognostics approach to nuclear component degradation modeling based on Gaussian Process Regression", \textit{Progress in Nuclear Energy}, Vol. 78, p. 141-154, \textit{2015}

\bibitem{b34} Datong Liu, Jingyue Pang, Jianbao Zhou, Yu Peng, Michael Pecht, "Prognostics for state of health estimation of lithium-ion batteries based on combination Gaussian process functional regression", \textit{Microelectronics Reliability},
Vol. 53, Issue 6, p. 832-839, \textit{2013}.

\bibitem{b35} Hong Sheng, Zhou Zheng, Lu Chen, Wang Baoqing, Zhao Tingdi, "Bearing remaining life prediction using Gaussian process regression with composite kernel functions". \textit{Journal of Vibroengineering}, Vol. 17, Issue 2, p. 695-704, \textit{2015}.

\bibitem{b36b}Li, Lingling, et al. "Remaining useful life prediction for lithium-ion batteries based on Gaussian processes mixture." PloS one 11.9 (2016): e0163004.

\bibitem{b36c}Datong Liu, Jingyue Pang, Jianbao Zhou and Yu Peng, "Data-driven prognostics for lithium-ion battery based on Gaussian Process Regression," Proceedings of the IEEE 2012 Prognostics and System Health Management Conference (PHM-2012 Beijing), Beijing, China, 2012, pp. 1-5, doi: 10.1109/PHM.2012.6228848.


\bibitem{b36} Biggio, Luca and Kastanis, Iason, "Prognostics and Health Management of Industrial Assets: Current Progress and Road Ahead", \textit{Frontiers in Artificial Intelligence}, Vol. 3, Issue Nov., p. 1-24, \textit{2020}

\bibitem{b37} Tipping, Michael E.. "Sparse Bayesian Learning and the Relevance Vector Machine." \textit{Journal of Machine Learning Research}  Vol. 1, p. 211--244, \textit{2001}.

\bibitem{b38} Rasmussen, Carl Edward. "Gaussian processes in machine learning." Summer school on machine learning. Springer, Berlin, Heidelberg, \textit{2003}.

\bibitem{b39} Y. Zheng, L. Wu, X. Li and C. Yin, "A relevance vector machine-based approach for remaining useful life prediction of power MOSFETs," \textit{Prognostics and System Health Management Conference (PHM-2014 Hunan), Zhangjiajie, China}, p. 642-646, \textit{2014}.

\bibitem{b40} Neal, Radford M. "MCMC using Hamiltonian dynamics." \textit{Handbook of markov chain monte carlo} 2.11 \textit{2011}.

\bibitem{ritter}Ritter, Hippolyt, Aleksandar Botev, and David Barber. "A scalable laplace approximation for neural networks." 6th International Conference on Learning Representations, ICLR 2018-Conference Track Proceedings. Vol. 6. International Conference on Representation Learning, 2018.

\bibitem{adams}Hernández-Lobato, José Miguel, and Ryan Adams. "Probabilistic backpropagation for scalable learning of bayesian neural networks." International Conference on Machine Learning. PMLR, 2015.

\bibitem{vi3}Teye, Mattias, Hossein Azizpour, and Kevin Smith. "Bayesian uncertainty estimation for batch normalized deep networks." International Conference on Machine Learning. PMLR, 2018.

\bibitem{b41} Samir Khan, Takehisa Yairi, "A review on the application of deep learning in system health management", \textit{Mechanical Systems and Signal Processing},
Vol. 107, p. 241-265, \textit{2018}.

\bibitem{42} G. Li, L. Yang, C. Lee, X. Wang and M. Rong, "A Bayesian Deep Learning RUL Framework Integrating Epistemic and Aleatoric Uncertainties," {IEEE Transactions on Industrial Electronics} \textit{2020}

\bibitem{b44} Benker, Maximilian and Furtner, Lukas and Semm, Thomas and Zaeh, Michael F, "Utilizing uncertainty information in remaining useful life estimation via Bayesian neural networks and Hamiltonian Monte Carlo", \textit{Journal of Manufacturing Systems}, \textit{2020}.

\bibitem{l1}Heimes, Felix O. "Recurrent neural networks for remaining useful life estimation." 2008 international conference on prognostics and health management. IEEE, 2008.

\bibitem{matlab}Rasmussen, Carl Edward, and Hannes Nickisch. "Gaussian processes for machine learning (GPML) toolbox." The Journal of Machine Learning Research 11 (2010): 3011-3015.

\bibitem{l2} Saha, Bhaskar, Kai Goebel, and Jon Christophersen. "Comparison of prognostic algorithms for estimating remaining useful life of batteries." Transactions of the Institute of Measurement and Control 31.3-4 (2009): 293-308.

\bibitem{l3} Li, Xiang, Qian Ding, and Jian-Qiao Sun. "Remaining useful life estimation in prognostics using deep convolution neural networks." Reliability Engineering \& System Safety 172 (2018): 1-11.

\bibitem{l4}Huang, Hong-Zhong, et al. "Support vector machine based estimation of remaining useful life: current research status and future trends." Journal of Mechanical Science and Technology 29.1 (2015): 151-163.

\bibitem{l5} Wu, Yuting, et al. "Remaining useful life estimation of engineered systems using vanilla LSTM neural networks." Neurocomputing 275 (2018): 167-179.

\bibitem{l6}Deutsch, Jason, and David He. "Using deep learning-based approach to predict remaining useful life of rotating components." IEEE Transactions on Systems, Man, and Cybernetics: Systems 48.1 (2017): 11-20.

\bibitem{b45} Zhao, Zhibin and Wu, Jingyao and Wong, David and Sun, Chuang and Yan, Ruqiang, "Probabilistic Remaining Useful Life Prediction Based on Deep Convolutional Neural Network", \textit{SSRN Electronic Journal} \textit{2020}.

\bibitem{b46} Xiang Li, Qian Ding, Jian-Qiao Sun, "Remaining useful life estimation in prognostics using deep convolution neural networks", \textit{Reliability Engineering {\&} System Safety}, Vol. 172, p. 1-11, \textit{2018}.

\bibitem{b47} André Listou Ellefsen, Emil Bjørlykhaug, Vilmar Æsøy, Sergey Ushakov, Houxiang Zhang, "Remaining useful life predictions for turbofan engine degradation using semi-supervised deep architecture", \textit{Reliability Engineering {\&} System Safety},Vol. 183, p. 240-251, \textit{2019}

\bibitem{b48} Paulo Roberto de Oliveira da Costa, Alp Akçay, Yingqian Zhang, Uzay Kaymak, "Remaining useful lifetime prediction via deep domain adaptation", \textit{Reliability Engineering {\&} System Safety}, Vol. 195, \textit{2020}.

\bibitem{b49} Saxena, Abhinav, et al. "Metrics for offline evaluation of prognostic performance." \textit{International Journal of Prognostics and health management} 1.1 (2010): 4-23.

\bibitem{fink2020potential}Fink, Olga, et al. "Potential, challenges and future directions for deep learning in prognostics and health management applications." Engineering Applications of Artificial Intelligence 92 (2020): 103678.

\bibitem{ruiz2020novel}Ruiz-Tagle Palazuelos, Andres, Enrique López Droguett, and Rodrigo Pascual. "A novel deep capsule neural network for remaining useful life estimation." Proceedings of the Institution of Mechanical Engineers, Part O: Journal of Risk and Reliability 234.1 (2020): 151-167.


\bibitem{l7}Sankararaman, Shankar, and Kai Goebel. "Uncertainty in prognostics and systems health management." International Journal of Prognostics and Health Management 6.010 (2015).


\end{thebibliography}
\end{document}